\documentclass[english]{article}
\usepackage[T1]{fontenc}
\usepackage[latin9]{inputenc}
\usepackage{geometry}
\usepackage{amstext}
\usepackage{graphicx}
\usepackage[authoryear]{natbib}

\makeatletter

\providecommand{\tabularnewline}{\\}

\usepackage{multicol} 

\makeatother

\usepackage{babel}
\begin{document}
\title{One weird trick for parallelizing convolutional neural networks}
\author{Alex Krizhevsky\\Google Inc.\\{\tt akrizhevsky@google.com }}
\maketitle
\begin{abstract}
I present a new way to parallelize the training of convolutional neural
networks across multiple GPUs. The method scales significantly better
than all alternatives when applied to modern convolutional neural
networks. 
\end{abstract}
\begin{multicols}{2}

\section{Introduction}

This is meant to be a short note introducing a new way to parallelize
the training of convolutional neural networks with stochastic gradient
descent (SGD). I present two variants of the algorithm. The first
variant perfectly simulates the synchronous execution of SGD on one
core, while the second introduces an approximation such that it no
longer perfectly simulates SGD, but nonetheless works better in practice.

\section{Existing approaches}

Convolutional neural networks are big models trained on big datasets.
So there are two obvious ways to parallelize their training: 
\begin{itemize}
\item across the model dimension, where different workers train different
parts of the model, and
\item across the data dimension, where different workers train on different
data examples.
\end{itemize}
These are called model parallelism and data parallelism, respectively. 

In model parallelism, whenever the model part (subset of neuron activities)
trained by one worker requires output from a model part trained by
another worker, the two workers must synchronize. In contrast, in
data parallelism the workers must synchronize model parameters (or
parameter gradients) to ensure that they are training a consistent
model. 

In general, we should exploit all dimensions of parallelism. Neither
scheme is better than the other a priori. But the relative degrees
to which we exploit each scheme should be informed by model architecture.
In particular, model parallelism is efficient when the amount of computation
per neuron activity is high (because the neuron activity is the unit
being communicated), while data parallelism is efficient when the
amount of computation per weight is high (because the weight is the
unit being communicated).

Another factor affecting all of this is batch size. We can make data
parallelism arbitrarily efficient if we are willing to increase the
batch size (because the weight synchronization step is performed once
per batch). But very big batch sizes adversely affect the rate at
which SGD converges as well as the quality of the final solution.
So here I target batch sizes in the hundreds or possibly thousands
of examples.

\section{Some observations}

Modern convolutional neural nets consist of two types of layers with
rather different properties:
\begin{itemize}
\item Convolutional layers cumulatively contain about 90-95\% of the computation,
about 5\% of the parameters, and have large representations.
\item Fully-connected layers contain about 5-10\% of the computation, about
95\% of the parameters, and have small representations. 
\end{itemize}
Knowing this, it is natural to ask whether we should parallelize these
two in different ways. In particular, data parallelism appears attractive
for convolutional layers, while model parallelism appears attractive
for fully-connected layers. 

This is precisely what I'm proposing. In the remainder of this note
I will explain the scheme in more detail and also mention several
nice properties.

\section{The proposed algorithm\label{sec:The-proposed-algorithm}}

\begin{figure*}
\begin{centering}
\includegraphics[scale=0.45]{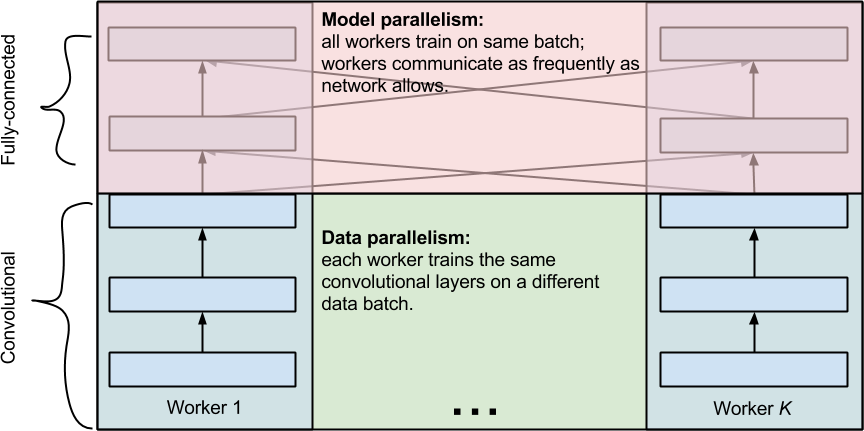}
\par\end{centering}

\caption{$K$ workers training a convolutional neural net with three convolutional
layers and two fully-connected layers. \label{fig:par-diagram}}

\end{figure*}

I propose that to parallelize the training of convolutional nets,
we rely heavily on data parallelism in the convolutional layers and
on model parallelism in the fully-connected layers. This is illustrated
in Figure \ref{fig:par-diagram} for $K$ workers. 

\begin{figure*}
\begin{centering}
\includegraphics[scale=0.55]{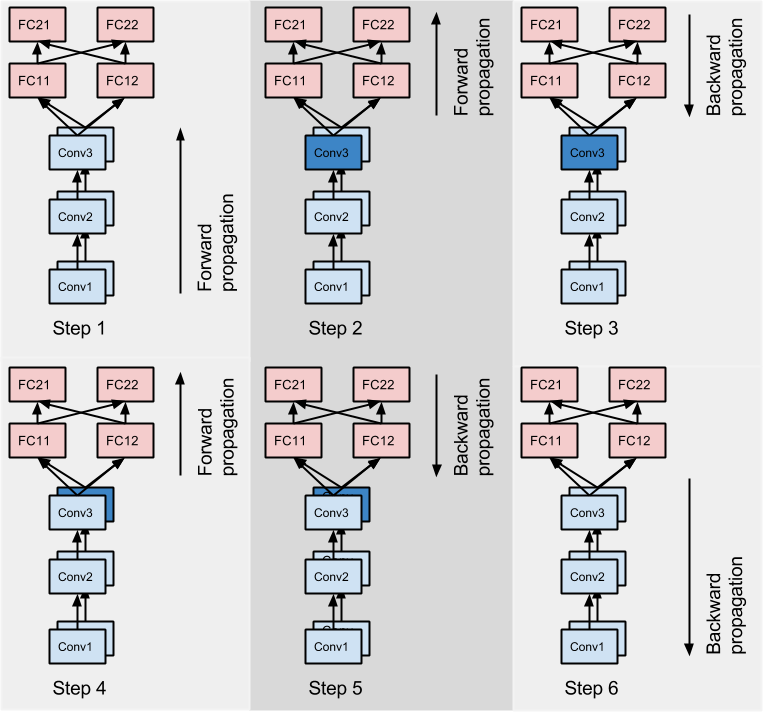}
\par\end{centering}

\caption{Illustration of the forward and backward propagations for scheme (b)
of Section \ref{sec:The-proposed-algorithm}, for $K=2$ workers.
Two-way data parallelism in the three convolutional layers is represented
with layers stacked on top of one another, while two-way model parallelism
in the two fully-connected layers is represented with layers laid
out next to one another. The standard two passes are replaced here
with six passes. Shading in the final convolutional layer indicates
the batch that is processed by the fully-connected layer. Note that,
as mentioned in Section \ref{sub:Variable-batch-size}, we are free
either to update the fully-connected weights during each of the backward
passes, or to accumulate a gradient and then update the entire net's
weights after the final backward pass. \label{fig:scheme-b-pic}}

\end{figure*}

In reference to the figure, the forward pass works like this:
\begin{enumerate}
\item Each of the $K$ workers is given a different data batch of (let's
say) 128 examples.
\item Each of the $K$ workers computes all of the convolutional layer activities
on its batch. 
\item To compute the fully-connected layer activities, the workers switch
to model parallelism. There are several ways to accomplish this:

\begin{enumerate}
\item Each worker sends its last-stage convolutional layer activities to
each other worker. The workers then assemble a big batch of activities
for $128K$ examples and compute the fully-connected activities on
this batch as usual.
\item One of the workers sends its last-stage convolutional layer activities
to all other workers. The workers then compute the fully-connected
activities on this batch of 128 examples and then begin to backpropagate
the gradients (more on this below) for these 128 examples. \textbf{In
parallel with this computation}, the next worker sends its last-stage
convolutional layer activities to all other workers. Then the workers
compute the fully-connected activities on this second batch of 128
examples, and so on.
\item All of the workers send $128/K$ of their last-stage convolutional
layer activities to all other workers. The workers then proceed as
in (b).
\end{enumerate}

It is worth thinking about the consequences of these three schemes.

\textbf{In scheme (a)}, all useful work has to pause while the big
batch of $128K$ images is assembled at each worker. Big batches also
consume lots of memory, and this may be undesirable if our workers
run on devices with limited memory (e.g. GPUs). On the other hand,
GPUs are typically able to operate on big batches more efficiently.

\textbf{In scheme (b)}, the workers essentially take turns broadcasting
their last-stage convolutional layer activities. The main consequence
of this is that much (i.e. $\frac{K-1}{K}$) of the communication
can be hidden -- it can be done in parallel with the computation of
the fully-connected layers. This seems fantastic, because this is
by far the most significant communication in the network.

\textbf{Scheme (c)} is very similar to scheme (b). Its one advantage
is that the communication-to-computation ratio is constant in $K$.
In schemes (a) and (b), it is proportional to $K.$ This is because
schemes (a) and (b) are always bottlenecked by the outbound bandwidth
of the worker that has to send data at a given ``step'', while scheme
(c) is able to utilize many workers for this task. This is a major
advantage for large $K$.

\end{enumerate}
The backward pass is quite similar:
\begin{enumerate}
\item The workers compute the gradients in the fully-connected layers in
the usual way. 
\item The next step depends on which of the three schemes was chosen in
the forward pass:

\begin{enumerate}
\item \textbf{In scheme (a)}, each worker has computed last-stage convolutional
layer activity gradients for the entire batch of $128K$ examples.
So each worker must send the gradient for each example to the worker
which generated that example in the forward pass. Then the backward
pass continues through the convolutional layers in the usual way.
\item \textbf{In scheme (b)}, each worker has computed the last-stage convolutional
layer activity gradients for one batch of 128 examples. Each worker
then sends these gradients to the worker which is responsible for
this batch of 128 examples. \textbf{In parallel with this}, the workers
compute the fully-connected forward pass on the next batch of 128
examples. After $K$ such forward-and-backward iterations through
the fully-connected layers, the workers propagate the gradients all
the way through the convolutional layers.
\item \textbf{Scheme (c)} is very similar to scheme (b). Each worker has
computed the last-stage convolutional layer activity gradients for
128 examples. This 128-example batch was assembled from $128/K$ examples
contributed by each worker, so to distribute the gradients correctly
we must reverse this operation. The rest proceeds as in scheme (b).
\end{enumerate}

I note again that, as in the forward pass, scheme (c) is the most
efficient of the three, for the same reasons.

\end{enumerate}
The forward and backward propagations for scheme (b) are illustrated
in Figure \ref{fig:scheme-b-pic} for the case of $K=2$ workers.

\subsection{Weight synchronization\label{sub:Weight-synchronization}}

Once the backward pass is complete, the workers can update the weights.
In the convolutional layers, the workers must also synchronize the
weights (or weight gradients) with one another. The simplest way that
I can think of doing this is the following:
\begin{enumerate}
\item Each worker is designated $1/K$th of the gradient matrix to synchronize. 
\item Each worker accumulates the corresponding $1/K$th of the gradient
from every other worker.
\item Each worker broadcasts this accumulated $1/K$th of the gradient to
every other worker.
\end{enumerate}
It's pretty hard to implement this step badly because there are so
few convolutional weights.

\subsection{Variable batch size\label{sub:Variable-batch-size}}

So what we have here in schemes (b) and (c) is a slight modification
to the standard forward-backward propagation which is, nonetheless,
completely equivalent to running synchronous SGD with a batch size
of $128K$. Notice also that schemes (b) and (c) perform $K$ forward
and backward passes through the fully-connected layers, each time
with a different batch of 128 examples. This means that we can, if
we wish, update the fully-connected weights\textbf{ }after each of
these partial backward passes, at virtually no extra computational
cost. We can think of this as using a batch size of 128 in the fully-connected
layers and $128K$ in the convolutional layers. With this kind of
variable batch size, the algorithm ceases to be a pure parallelization
of SGD, since it no longer computes a gradient update for any consistent
model in the convolutional layers. But it turns out that this doesn't
matter much in practice. As we take the effective batch size, $128K$,
into the thousands, using a smaller batch size in the fully-connected
layers leads to faster convergence to better minima.

\section{Experiments\label{sec:Experiments}}

The first question that I investigate is the accuracy cost of larger
batch sizes. This is a somewhat complicated question because the answer
is dataset-dependent. Small, relatively homogeneous datasets benefit
from smaller batch sizes more so than large, heterogeneous, noisy
datasets. Here, I report experiments on the widely-used ImageNet 2012
contest dataset (ILSVRC 2012) \citep{deng2009imagenet}. At 1.2 million
images in 1000 categories, it falls somewhere in between the two extremes.
It isn't tiny, but it isn't ``internet-scale'' either. With current
GPUs (and CPUs) we can afford to iterate through it many times when
training a model. 

The model that I consider is a minor variation on the winning model
from the ILSVRC 2012 contest \citep{krizhevsky2012imagenet}. The
main difference is that it consists of one ``tower'' instead of
two. This model has 0.2\% more parameters and 2.4\% fewer connections
than the two-tower model. It has the same number of layers as the
two-tower model, and the $(x,y)$ map dimensions in each layer are
equivalent to the $(x,y)$ map dimensions in the two-tower model.
The minor difference in parameters and connections arises from a necessary
adjustment in the number of kernels in the convolutional layers, due
to the unrestricted layer-to-layer connectivity in the single-tower
model.%
\footnote{In detail, the single-column model has 64, 192, 384, 384, 256 filters
in the five convolutional layers, respectively. %
} Another difference is that instead of a softmax final layer with
multinomial logistic regression cost, this model's final layer has
1000 independent logistic units, trained to minimize cross-entropy.
This cost function performs equivalently to multinomial logistic regression
but it is easier to parallelize, because it does not require a normalization
across classes.%
\footnote{This is not an important point with only 1000 classes. But with tens
of thousands of classes, the cost of normalization becomes noticeable. %
} I trained all models for exactly 90 epochs, and multiplied the learning
rate by $250^{-1/3}$ at 25\%, 50\%, and 75\% training progress.

The weight update rule that I used was

\begin{eqnarray*}
\Delta w & := & \mu\Delta w+\epsilon\left(\langle\frac{\partial E}{\partial w}\rangle_{i}-\omega w\right)\\
w & := & w+\Delta w
\end{eqnarray*}
where $\mu$ is the coefficient of momentum, $\omega$ is the coefficient
of weight decay, $\epsilon$ is the learning rate, and $\langle\frac{\partial E}{\partial w}\rangle_{i}$
denotes the expectation of the weight gradient for a batch $i$. 

When experimenting with different batch sizes, one must decide how
to adjust the hyperparameters $\mu,\omega,$ and $\epsilon$. It seems
plausible that the smoothing effects of momentum may be less necessary
with bigger batch sizes, but in my experiments I used $\mu=0.9$ for
all batch sizes. Theory suggests that when multiplying the batch size
by $k$, one should multiply the learning rate $\epsilon$ by $\sqrt{k}$
to keep the variance in the gradient expectation constant. How should
we adjust the weight decay $\omega$? Given old batch size $N$ and
new batch size $N^{\prime}=k\cdot N$, we'd like to keep the total
weight decay penalty constant. Note that with batch size $N$, we
apply the weight decay penalty $k$ times more frequently than we
do with batch size $N^{\prime}$. So we'd like $k$ applications of
the weight decay penalty under batch size $N$ to have the same effect
as one application of the weight decay penalty under batch size $N^{\prime}$.
Assuming $\mu=0$ for now, $k$ applications of the weight decay penalty
under batch size $N$, learning rate $\epsilon$, and weight decay
coefficient $\omega$ give
\begin{eqnarray*}
w_{k} & = & w_{k-1}-\epsilon\omega w_{k-1}\\
 & = & w_{k-1}(1-\epsilon\omega)\\
 & = & w_{0}(1-\epsilon\omega)^{k}.
\end{eqnarray*}
While one application of weight decay under batch size $N^{\prime}$,
learning rate $\epsilon^{\prime}$ and weight decay coefficient $\omega^{\prime}$
gives 
\begin{eqnarray*}
w_{1}^{\prime} & = & w_{0}-\epsilon^{\prime}\omega^{\prime}w_{0}\\
 & = & w_{0}(1-\epsilon^{\prime}\omega^{\prime})
\end{eqnarray*}
so we want to pick $\omega^{\prime}$ such that 
\[
(1-\epsilon\omega)^{k}=1-\epsilon^{\prime}\omega^{\prime}
\]
which gives 
\begin{eqnarray*}
\omega^{\prime} & = & \frac{1}{\epsilon^{\prime}}\cdot\left(1-\left(1-\epsilon\omega\right){}^{k}\right)\\
 & = & \frac{1}{\sqrt{k}\epsilon}\cdot\left(1-\left(1-\epsilon\omega\right){}^{k}\right).
\end{eqnarray*}
So, for example, if we trained a net with batch size $N=128$ and
$\epsilon=0.01,\omega=0.0005$, the theory suggests that for batch
size $N^{\prime}=1024$ we should use $\epsilon^{\prime}=\sqrt{8}\cdot0.01$
and $\omega^{\prime}\approx0.0014141888$. Note that, as $\epsilon\to0$,
$\omega^{\prime}=\frac{1}{\sqrt{k}\epsilon}\cdot\left(1-\left(1-\epsilon\omega\right){}^{k}\right)\to\sqrt{k}\cdot\omega$,
an easy approximation which works for the typical $\epsilon$s used
in neural nets. In our case, the approximation yields $\omega^{\prime}\approx\sqrt{8}\cdot\omega\approx0.0014142136$.
The acceleration obtained due to momentum $\mu=0.9$ is no greater
than that obtained by multiplying $\epsilon$ by 10, so the $\sqrt{8}\cdot\omega$
approximation remains very accurate.

Theory aside, for the batch sizes considered in this note, the heuristic
that I found to work the best was to multiply the learning rate by
$k$ when multiplying the batch size by $k.$ I can't explain this
discrepancy between theory and practice%
\footnote{This heuristic does eventually break down for batch sizes larger than
the ones considered in this note.%
}. Since I multiplied the learning rate $\epsilon$ by $k$ instead
of $\sqrt{k}$, and the total weight decay coefficient is $\epsilon^{\prime}\omega^{\prime}$,
I used $\omega^{\prime}=\omega=0.0005$ for all experiments. 

As in \citep{krizhevsky2012imagenet}, I trained on random $224\times224$
patches extracted from $256\times256$ images, as well as their horizontal
reflections. I computed the validation error from the center $224\times224$
patch.

The machine on which I performed the experiments has eight NVIDIA
K20 GPUs and two Intel 12-core CPUs. Each CPU provides two PCI-Express
2.0 lanes for four GPUs. GPUs which have the same CPU ``parent''
can communicate amongst themselves simultaneously at the full PCI-Express
2.0 rate (about 6GB/sec) through a PCI-Express switch. Communication
outside this set must happen through the host memory and incurs a
latency penalty, as well as a throughput penalty of 50\% if all GPUs
wish to communicate simultaneously.

\subsection{Results\label{sub:Results}}

Table \ref{tab:error-rates} summarizes the error rates and training
times of this model using scheme (b) of Section \ref{sec:The-proposed-algorithm}.
The main take-away is that there is an accuracy cost associated with
bigger batch sizes, but it can be greatly reduced by using the variable
batch size trick described in Section \ref{sub:Variable-batch-size}.
The parallelization scheme scales pretty well for the model considered
here, but the scaling is not quite linear. Here are some reasons for
this:
\begin{itemize}
\item The network has three dense matrix multiplications near the output.
Parallel dense matrix multiplication is quite inefficient for the
matrix sizes used in this network. With 6GB/s PCI-Express links and
2 TFLOP GPUs, more time is spent communicating than computing the
matrix products for $4096\times4096$ matrices.%
\footnote{Per sample, each GPU must perform $4096\times512\times2$ FLOPs, which
takes $2.09\mu\textrm{s}$ at 2 TFLOPs/sec, and each GPU must receive
$4096$ floats, which takes $2.73\mu\mathrm{s}$ at 6GB/sec.%
} We can expect better scaling if we increase the sizes of the matrices,
or replace the dense connectivity of the last two hidden layers with
some kind of restricted connectivity. 
\item One-to-all broadcast/reduction of scheme (b) is starting to show its
cost. Scheme (c), or some hybrid between scheme (b) and scheme (c),
should be better.
\item Our 8-GPU machine does not permit simultaneous full-speed communication
between all 8 GPUs, but it does permit simultaneous full-speed communication
between certain subsets of 4 GPUs. This particularly hurts scaling
from 4 to 8 GPUs.
\end{itemize}
\begin{table*}
\begin{centering}
\begin{tabular}{|c|c|c|c|c|c|}
\hline 
\textbf{GPUs} & \textbf{Batch size} & \textbf{Cross-entropy} & \textbf{Top-1 error} & \textbf{Time} & \textbf{Speedup}\tabularnewline
\hline 
\hline 
1 & $(128,128)$ & 2.611 & 42.33\% & 98.05h & 1x\tabularnewline
\hline 
\hline 
2 & $(256,256)$ & 2.624 & 42.63\% & 50.24h & 1.95x\tabularnewline
\hline 
2 & $(256,128)$ & 2.614 & 42.27\% & 50.90h & 1.93x\tabularnewline
\hline 
\hline 
4 & $(512,512)$ & 2.637 & 42.59\% & 26.20h & 3.74x\tabularnewline
\hline 
4 & $(512,128)$ & 2.625 & 42.44\% & 26.78h & 3.66x\tabularnewline
\hline 
\hline 
\multicolumn{1}{|c|}{8} & $(1024,1024)$ & 2.678 & 43.28\% & 15.68h & 6.25x\tabularnewline
\hline 
8 & $(1024,128)$ & 2.651 & 42.86\% & 15.91h & 6.16x\tabularnewline
\hline 
\end{tabular}
\par\end{centering}

\caption{Error rates on the validation set of ILSVRC 2012, with the model described
in Section \ref{sec:Experiments}. \emph{Batch size} $(m,n)$ indicates
an effective batch size of $m$ in the convolutional layers and $n$
in the fully-connected layers. All models use data parallelism in
the convolutional layers and model parallelism in the fully-connected
layers. \emph{Time} indicates total training time in hours.\label{tab:error-rates} }
\end{table*}

\section{Comparisons to other work on parallel convolutional neural network
training}

The results of Table \ref{tab:error-rates} compare favorably to published
alternatives. In \citep{yadan2013multi}, the authors parallelize
the training of the convolutional neural net from \citep{krizhevsky2012imagenet}
using model parallelism and data parallelism, but they use the same
form of parallelism in every layer. They achieved a speedup of 2.2x
on 4 GPUs, relative to a 1-GPU implementation that takes 226.8 hours
to train for 90 epochs on an NVIDIA GeForce Titan. In \citep{paine2013gpu},
the authors implement asynchronous SGD \citep{niu2011hogwild,dean2012large}
on a GPU cluster with fast interconnects and use it to train the convolutional
neural net of \citep{krizhevsky2012imagenet} using model parallelism
and data parallelism. They achieved a speedup of 3.2x on 8 GPUs, relative
to a 1-GPU implementation that takes 256.8 hours to train on an NVIDIA
K20X. Furthermore, this 3.2x speedup came at a rather significant
accuracy cost: their 8-GPU model achieved a final validation error
rate of 45\%.

\section{Other work on parallel neural network training}

In \citep{coates2013deep}, the authors use a GPU cluster to train
a locally-connected neural network on images. To parallelize training,
they exploit the fact that their network is locally-connected but
not convolutional. This allows them to distribute workers spatially
across the image, and only neuron activations near the edges of the
workers' areas of responsibility need to be communicated. This scheme
could potentially work for convolutional nets as well, but the convolutional
weights would need to be synchronized amongst the workers as well.
This is probably not a significant handicap as there aren't many convolutional
weights. The two other disadvantages of this approach are that it
requires synchronization at every convolutional layer, and that with
8 or more workers, each worker is left with a rather small area of
responsibility (particularly near the upper layers of the convolutional
net), which has the potential to make computation inefficient. Nonetheless,
this remains an attractive dimension of parallelization for convolutional
neural nets, to be exploited alongside the other dimensions. 

The work of \citep{coates2013deep} extends the work of \citep{dean2012large},
which introduced this particular form of model parallelism for training
a locally-connected neural network. This work also introduced the
version of the asynchronous SGD algorithm employed by \citep{paine2013gpu}.
Both of these works are in turn based on the work of \citep{niu2011hogwild}
which introduced asynchronous SGD and demonstrated its efficacy for
models with sparse gradients.

\section{Conclusion}

The scheme introduced in this note seems like a reasonable way to
parallelize the training of convolutional neural networks. The fact
that it works quite well on existing model architectures, which have
not been adapted in any way to the multi-GPU setting, is promising.
When we begin to consider architectures which are more suited to the
multi-GPU setting, we can expect even better scaling. In particular,
as we scale the algorithm past 8 GPUs, we should:
\begin{itemize}
\item Consider architectures with some sort of restricted connectivity in
the upper layers, in place of the dense connectivity in current nets.
We might also consider architectures in which a fully-connected layer
on one GPU communicates only a small, linear projection of its activations
to other GPUs. 
\item Switch from scheme (b) to scheme (c) of Section \ref{sec:The-proposed-algorithm},
or some hybrid between schemes (b) and (c). 
\item Reduce the effective batch size by using some form of restricted model
parallelism in the convolutional layers, as in the two-column network
of \citep{krizhevsky2012imagenet}. 
\end{itemize}
We can expect some loss of accuracy when training with bigger batch
sizes. The magnitude of this loss is dataset-dependent, and it is
generally smaller for larger, more varied datasets. 

\bibliographystyle{plainnat}
\bibliography{sgd-note}

\end{multicols}
\end{document}